\documentclass[final,5p,times,twocolumn]{elsarticle}

\usepackage{graphicx}
\usepackage{amsmath}
\usepackage{amssymb}
\usepackage{pifont}
\usepackage{makecell}
\usepackage{array}
\usepackage{color}
\usepackage{tabularx}

\journal{Journal of Agriculture and Food Research}

\begin{document}

\begin{frontmatter}

\title{Agri-Sim: Agricultural Simulation Platform for Embodied Intelligence Evaluation in Greenhouse Robotics}

\author[aff1]{Shuhan Shi}
\author[aff1]{Zhenfeng Xue\corref{cor1}}
\ead{zfxue0903@shu.edu.cn}
\author[aff2]{Minghao Mei}
\author[aff3]{Chao Zheng}
\author[aff1]{Nan Li}
\author[aff1]{Zhonghua Miao\corref{cor1}}
\ead{zhhmiao@shu.edu.cn}

\cortext[cor1]{Corresponding authors.}
\address[aff1]{School of Mechatronic Engineering and Automation, Shanghai University, Shanghai 200444, China}
\address[aff2]{Sino-European School of Technology of Shanghai University, Shanghai University, Shanghai 200444, China}
\address[aff3]{School of Future Technology (Institute of Artificial Intelligence), Shanghai University, Shanghai 200444, China}

\begin{abstract}
Agricultural-robot development requires simulation environments that can jointly support realistic scene construction, virtual sensing, autonomous navigation, motion planning, and manipulation-task execution. This paper presents Agri-Sim, a Unity and ROS2-based simulation platform for the closed-loop development and functional evaluation of agricultural robots. The platform contains a configurable tomato-greenhouse environment, a mobile dual-arm harvesting robot, virtual RGB-D, LiDAR, IMU, and joint sensors, and a bidirectional communication interface between Unity and ROS2. Unity is responsible for scene rendering, rigid-body dynamics, collision detection, virtual sensing, and task-state execution, whereas ROS2 and MoveIt~2 provide localization, navigation, collision-aware motion planning, inverse kinematics, and trajectory generation. Autonomous greenhouse navigation and dual-arm tomato harvesting were used to evaluate the complete simulation workflow. The experiments covered virtual sensor publication, ROS2-based navigation, collision-aware motion planning, mobile-base control, tomato acquisition, inter-arm handover, and box placement. The results demonstrate that Agri-Sim supports closed-loop integration and repeatable functional evaluation of navigation and manipulation workflows in a controlled virtual greenhouse, providing a practical foundation for subsequent algorithm development and Sim-to-Real studies.
\end{abstract}

\end{frontmatter}

\section{Introduction}
\label{sec:introduction}

Agricultural robots have been increasingly investigated for environmental monitoring, autonomous navigation, crop inspection, spraying, phenotyping, and selective harvesting in response to labor shortages and the growing demand for efficient agricultural production \cite{oliveira2021agriculture,droukas2023survey,bagagiolo2022greenhouse}. Among these applications, robotic harvesting remains particularly challenging because a robot must navigate through confined crop rows, perceive partially occluded crops, estimate target poses, and execute
manipulation motions without damaging plants or greenhouse facilities. Crop deformability, irregular geometry, and biological variability further increase the difficulty of these tasks \cite{bac2014harvesting,droukas2023survey}. Consequently, greenhouse automation requires the coordinated operation of perception, localization, navigation, manipulation, and control \cite{bagagiolo2022greenhouse,nasirahmadi2022toward}.

\begin{figure}[h]
    \centering
    \includegraphics[width=\columnwidth]
    {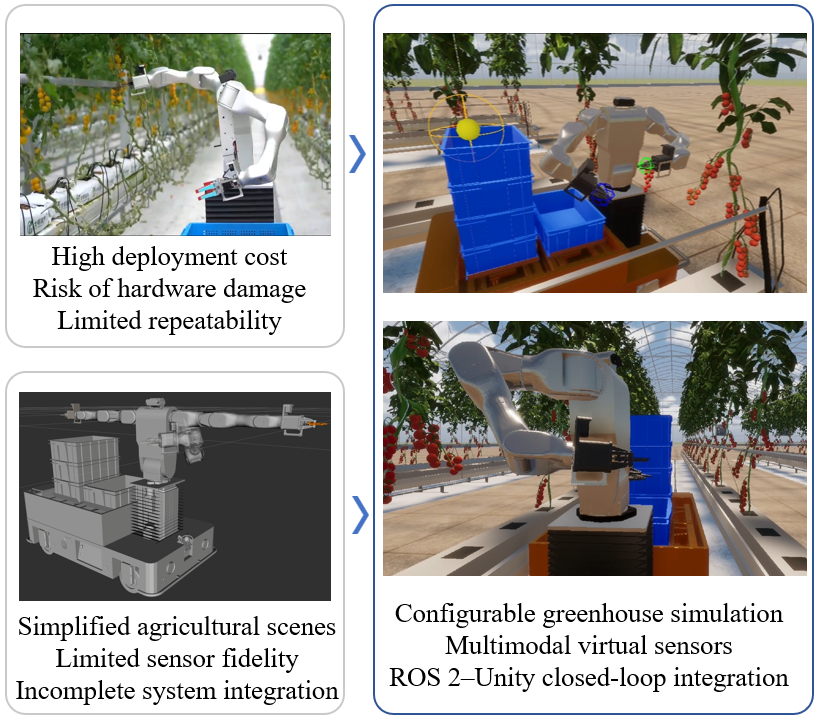}
    \caption{Motivation for Agri-Sim. The platform bridges the gap between
    costly and difficult-to-repeat real-world development and conventional
    simulation with limited agricultural-scene, sensor, and system fidelity.}
    \label{fig:sim_to_real_gap}
\end{figure}

Direct experiments with physical agricultural robots are costly, time-consuming, and difficult to repeat because illumination, crop growth, fruit arrangement, and obstacle distribution vary over time. Simulation allows robot, sensor, and environmental configurations to be controlled before physical deployment. General-purpose simulators, including Gazebo, Webots, and CoppeliaSim, have been widely used for robot modeling, sensing, navigation, and manipulation \cite{koenig2004gazebo,michel2004webots,rohmer2013vrep}. However, simulators differ considerably in rendering, physical dynamics, sensor modeling, software interfaces, and scalability \cite{collins2021review}. Agricultural applications additionally require dense vegetation, crop occlusion, illumination variation, and interactions among mobile platforms, manipulators, plants, and greenhouse structures. Modern game engines provide flexible tools for constructing detailed and programmable virtual environments. Unity supports physically based rendering, configurable illumination, interactive scene construction, and interfaces for intelligent-agent development \cite{juliani2020mlagents}. However, it does not independently provide the complete localization, navigation, motion-planning, and control capabilities required by an integrated robotic system. ROS provides a modular framework for robot perception, planning, and control \cite{macenski2022ros2}, while MoveIt supports robot-state management, inverse kinematics, collision checking, motion planning, and trajectory execution \cite{coleman2014moveit}. Combining Unity with ROS therefore enables visually detailed agricultural simulation to interact with established robotic software.

Unity--ROS2 communication has been investigated in previous robotic simulation studies \cite{babaians2018ros2unity3d}, while agricultural robotics research has separately addressed greenhouse navigation, crop-row navigation, mobile-robot simulation, and harvesting manipulation
\cite{guyonneau2022lidar,iqbal2020simulation,li2023hybrid,zheng2024tomato}. Nevertheless, many existing systems focus on an individual task or subsystem. A unified platform in which a mobile base and harvesting manipulators share the same greenhouse environment, multimodal sensor observations, coordinate frames, robot-state feedback, and control interfaces remains necessary for system-level development and evaluation. As illustrated in Fig.~\ref{fig:sim_to_real_gap}, such a platform should combine the safety and repeatability of simulation with the scene, sensor, and communication fidelity required by agricultural robot tasks.

To address these requirements, this paper presents Agri-Sim, a configurable ROS2--Unity simulation platform for integrated agricultural robot development and evaluation. The platform combines a virtual tomato greenhouse, a mobile dual-arm harvesting robot, and configurable RGB-D, LiDAR, IMU, and joint-state sensors. Weather, illumination, scene arrangement, sensor configuration, and simulation time are adjustable.

Agri-Sim establishes bidirectional communication between Unity and ROS2 for multimodal sensor observations, robot states, target poses, mobile-base commands, gripper commands, and manipulator trajectories. Unity performs environment rendering, physics simulation, virtual-sensor generation, and robot-motion execution, whereas ROS2 and MoveIt~2 provide coordinate transformation, navigation, collision checking, motion planning, and task-level control. Autonomous greenhouse navigation and a complete dual-arm tomato-harvesting workflow verify integrated communication, planning, and execution rather than introduce new algorithms.

The main contributions of this work are summarized as follows:

\begin{itemize}

    \item We develop Agri-Sim, a configurable ROS2--Unity platform that integrates a tomato-greenhouse environment, a mobile dual-arm robot, and multimodal virtual sensors.

    \item We provide adjustable environmental, sensor, scene, and simulation-time parameters for controlled and repeatable agricultural robot experiments.

    \item We establish bidirectional ROS2--Unity communication for sensor data, robot-state feedback, target poses, base and gripper commands, and time-parameterized manipulator trajectories.

    \item We validate the complete closed-loop workflow through autonomous greenhouse navigation and a dual-arm harvesting task comprising acquisition, inter-arm handover, and box placement.

\end{itemize}

\section{Related Work}
\label{sec:related_work}

\subsection{Robot Simulation Platforms}
\label{subsec:general_simulators}

Robot simulators provide controlled and repeatable environments for robot modeling, virtual sensing, algorithm development, and system-level evaluation. Existing platforms differ in physical fidelity, rendering quality, sensor modeling, middleware integration, computational
performance, and environment-construction capabilities \cite{collins2021review}. Gazebo is widely used for rigid-body dynamics, sensor simulation, and ROS-based evaluation \cite{koenig2004gazebo}.
Webots provides predefined robot and sensor models with multilingual programming interfaces \cite{michel2004webots}, while CoppeliaSim supports distributed control through embedded scripts, remote APIs, and external nodes \cite{rohmer2013vrep}. Although these platforms support
mobile robots and manipulators, constructing detailed crop models, greenhouse structures, and harvesting scenarios generally requires considerable additional development.

MuJoCo is commonly used for contact-rich control and robot learning \cite{todorov2012mujoco}. Isaac Gym further supports large-scale GPU-accelerated robot-learning experiments \cite{makoviychuk2021isaacgym}. Its main strengths are dynamics simulation and parallel learning rather than ready-to-use agricultural environments or integrated greenhouse navigation and harvesting workflows.

Game engines provide flexible tools for developing visually detailed and programmable environments. Unity supports physically based rendering, configurable illumination, programmable cameras, and interactive scene construction, while Unity ML-Agents provides interfaces for intelligent-agent development \cite{juliani2020mlagents}. However, Unity alone does not supply the complete localization, navigation, motion-planning, and control stack required by an
integrated robotic system. 

ROS provides reusable components for perception, planning, control, and interprocess communication \cite{macenski2022ros2}, while MoveIt supports robot-state management, inverse kinematics, collision checking, motion planning, and trajectory execution \cite{coleman2014moveit}. Previous studies have demonstrated communication between Unity and ROS
\cite{babaians2018ros2unity3d}. Extending this connection to greenhouse mobile manipulation nevertheless requires consistent coordinate frames, timestamps, sensor messages, joint states, base commands, gripper commands, and manipulator trajectories.

Simulation can also facilitate the transfer of learned methods to physical systems. Domain randomization varies visual properties such as illumination, textures, and camera parameters \cite{tobin2017domain}, whereas dynamics randomization varies mass, friction, damping, and
actuator characteristics \cite{peng2018sim}. Agri-Sim exposes configurable environmental and simulation parameters that can support such studies in the future; however, reinforcement-learning training and Sim-to-Real transfer are outside the experimental scope of this work.

\begin{table*}[t]
    \centering
    \caption{Comparison of representative agricultural robot simulation systems and the proposed Agri-Sim platform. ``Partial'' indicates that the capability is limited to a subsystem or is not demonstrated as part of a complete closed-loop workflow.}
    \label{tab:related_work_comparison}
    \footnotesize
    \setlength{\tabcolsep}{3.5pt}
    \renewcommand{\arraystretch}{1.18}
    \resizebox{\textwidth}{!}{%
    \begin{tabular}{lccccccc}
        \hline
        \textbf{System or study} &
        \textbf{Primary platform} &
        \textbf{Agricultural scene} &
        \textbf{Multimodal sensing} &
        \textbf{Ground navigation} &
        \textbf{Manipulation} &
        \textbf{ROS integration} &
        \textbf{Unified closed loop} \\
        \hline

        Iqbal \textit{et al.} \cite{iqbal2020simulation} &
        Robot simulator &
        Yes &
        Partial &
        Yes &
        -- &
        Partial &
        -- \\

        Ivanovic \textit{et al.} \cite{ivanovic2022render} &
        Gazebo--Blender &
        Yes &
        Yes &
        -- &
        -- &
        Yes &
        Partial \\

        Agri-fly \cite{zha2024agrifly} &
        Custom UAV simulator &
        Yes &
        Yes &
        -- &
        -- &
        Partial &
        Partial \\

        Gazebo Plants \cite{deng2024gazeboplants} &
        Gazebo &
        Yes &
        -- &
        -- &
        Yes &
        Yes &
        Partial \\

        Li and Xiang \cite{li2024photorealistic} &
        Unreal Engine--ROS &
        Yes &
        Partial &
        Partial &
        -- &
        Yes &
        Partial \\

        DigiHortiRobot \cite{fernandez2025digihortirobot} &
        Digital-twin framework &
        Yes &
        Yes &
        Partial &
        Yes &
        Yes &
        Partial \\

        \textbf{Agri-Sim (ours)} &
        \textbf{Unity--ROS2} &
        \textbf{Yes} &
        \textbf{Yes} &
        \textbf{Yes} &
        \textbf{Yes} &
        \textbf{Yes} &
        \textbf{Yes} \\

        \hline
    \end{tabular}%
    }
\end{table*}

\subsection{Simulation for Agricultural Robotics}
\label{subsec:agricultural_simulation}

Agricultural robots have been developed for crop monitoring, navigation, spraying, phenotyping, transportation, and harvesting \cite{oliveira2021agriculture,droukas2023survey}. Greenhouse harvesting is particularly difficult because of confined spaces, crop occlusion, illumination variation, irregular plant geometry, and the need for safe robot--plant interaction
\cite{bac2014harvesting,bagagiolo2022greenhouse}. Simulation provides a safe and repeatable means of developing these systems before physical deployment.

For agricultural navigation, Iqbal \textit{et al.} developed a simulated autonomous robot for LiDAR-based in-field phenotyping and navigation \cite{iqbal2020simulation}. Guyonneau \textit{et al.} investigated LiDAR-only navigation in structured crop environments \cite{guyonneau2022lidar}, while Martini \textit{et al.} studied synthetic-data generation for visual navigation in row crops \cite{martini2024synthetic}. These studies demonstrate the value of simulated geometric and visual observations, but primarily focus on navigation or perception rather than integrated mobile manipulation.

Other platforms address specific robot types or agricultural capabilities. Ivanovic \textit{et al.} combined Gazebo and Blender Cycles for aerial crop monitoring and yield estimation
\cite{ivanovic2022render}. Agri-fly provides dynamics, sensing, plant models, and autonomous-flight functions for agricultural UAVs \cite{zha2024agrifly}. Gazebo Plants models plant deformation during robot contact using Cosserat rods \cite{deng2024gazeboplants}, whereas
Li and Xiang developed a photorealistic Unreal Engine--ROS environment for image generation and robot-position verification \cite{li2024photorealistic}. These systems provide specialized
capabilities in aerial simulation, plant mechanics, or visual realism, but do not primarily target a complete ground-based greenhouse navigation and harvesting workflow.

Harvesting research has also examined individual perception, planning, and control components. Li \textit{et al.} proposed hybrid visual-servo control for cherry-tomato harvesting \cite{li2023hybrid}, and Zheng \textit{et al.} investigated fruit-direction recognition and nested
grasping strategies \cite{zheng2024tomato}. Digital-twin research provides a broader framework for connecting environments, robot states, sensor data, and decision-making modules
\cite{nasirahmadi2022toward}. DigiHortiRobot, for example, introduced an AI-driven digital-twin architecture for hydroponic horticulture and dual-arm automation \cite{fernandez2025digihortirobot}. Nevertheless, an integrated ROS2--Unity implementation that connects multimodal sensing, mobile navigation, MoveIt~2 planning, and trajectory execution
within the same greenhouse environment remains valuable.

Table~\ref{tab:related_work_comparison} compares representative agricultural simulation studies according to their reported scope. The comparison is intended to show differences in system integration rather than rank the physical or visual fidelity of the platforms.

\subsection{Research Gap and Positioning of Agri-Sim}
\label{subsec:research_gap}

Existing agricultural simulation studies have advanced synthetic-data generation, crop-row navigation, aerial-robot simulation, deformable-plant modeling, photorealistic rendering, digital twins, and harvesting manipulation. However, many systems are designed for a particular robot type, sensing modality, or isolated task. Multimodal sensor generation, ROS-side algorithms, robot-state feedback, mobile-base control, target-pose exchange, manipulator planning, and trajectory execution are therefore not always connected within the same agricultural simulation environment.

Agri-Sim addresses this system-integration gap by combining a Unity-based tomato greenhouse, a mobile dual-arm harvesting robot, configurable RGB-D, LiDAR, IMU, and joint-state sensors, and bidirectional ROS2 communication. Unity performs environment rendering, physics simulation, virtual-sensor generation, parameter configuration, and robot-motion execution. ROS2 and MoveIt~2 provide coordinate transformation, navigation, collision checking, motion planning, and task-level control.

Agri-Sim is not intended to replace specialized platforms for high-accuracy plant deformation, biological growth modeling, or massively parallel reinforcement-learning training. Instead, it
provides a unified and repeatable workflow in which greenhouse navigation and tomato manipulation share the same robot model, environment, sensor interfaces, communication framework, and execution loop. Its configurable environmental parameters, control interfaces, collision events, and simulation-time mechanisms also provide a basis for future learning-based and Sim-to-Real studies.

\section{Proposed Platform}
\label{sec:platform}

Agri-Sim is a Unity--ROS2 simulation platform developed for the integrated evaluation of agricultural mobile manipulation. It combines a configurable tomato-greenhouse environment, a mobile dual-arm harvesting robot, multimodal virtual sensors, and bidirectional communication with ROS2 and MoveIt~2. Unity performs scene rendering, physics simulation, sensor generation, collision detection, and robot-motion execution, whereas ROS2 provides perception, navigation, coordinate transformation, motion planning, and task-level control.

Unlike simulation environments designed for an isolated perception or control component, Agri-Sim connects sensing, planning, control, and execution within a common robot and greenhouse instance. Observations and robot states generated in Unity are transmitted to ROS2. The resulting base commands, gripper commands, and joint trajectories are returned to Unity for execution. The updated observations are subsequently republished, thereby forming a closed simulation loop.

\subsection{System Architecture}
\label{subsec:architecture}

\begin{figure*}[t]
    \centering
    \includegraphics[width=0.98\textwidth]{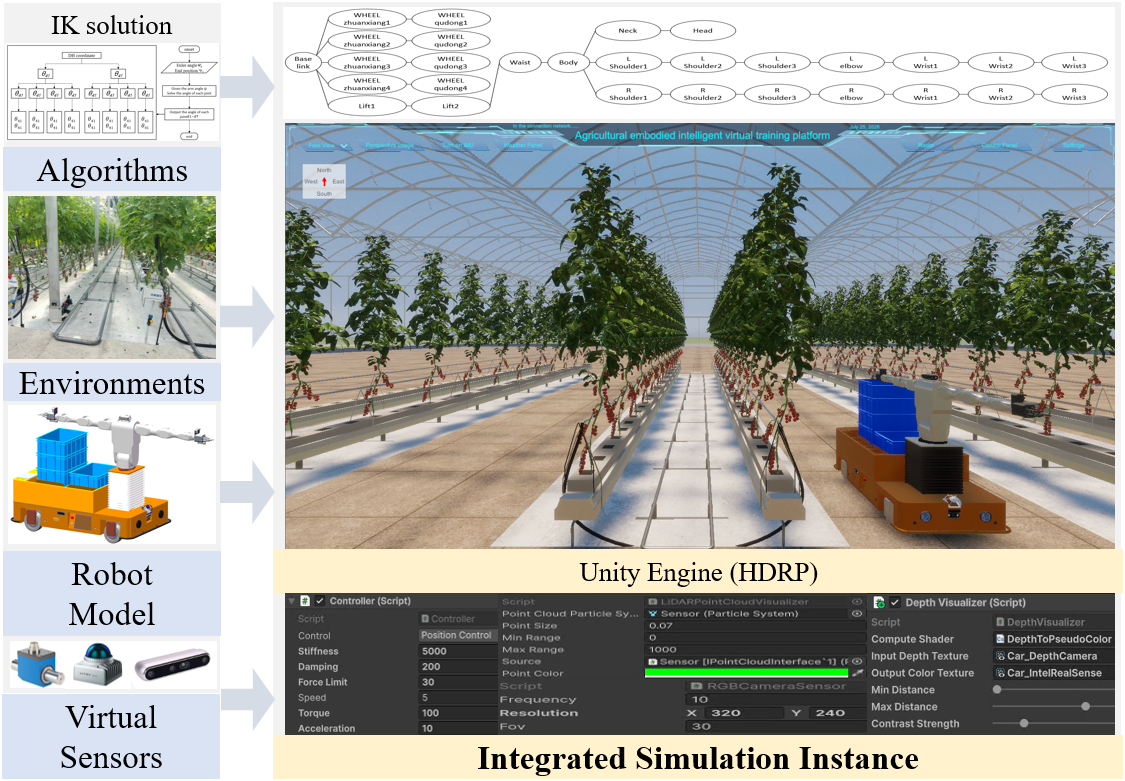}
    \caption{Core components of Agri-Sim and their integration in the
    Unity HDRP greenhouse simulation.}
    \label{fig:core_modules}
\end{figure*}

The core components of Agri-Sim are shown in Fig.~\ref{fig:core_modules}. Greenhouse assets, robot models, sensing modules, and algorithm interfaces are assembled in Unity HDRP to construct
a configurable agricultural simulation scenario.

The platform consists of four functional layers. The environment layer contains the greenhouse structure, crops, cultivation facilities, illumination, weather effects, and obstacles. The robot-and-sensor layer contains the mobile dual-arm robot, its articulation and collision models, and the RGB-D, LiDAR, IMU, and robot-state sensors. The communication layer converts Unity data into ROS2 messages and transfers ROS2 control commands back to Unity. The algorithm layer contains ROS2 perception and navigation nodes together with MoveIt~2 coordinate-transformation, collision-checking, inverse-kinematics, and motion-planning modules.

The closed-loop relationship between these layers is represented as

\begin{equation}
\mathcal{E}_{t}
\rightarrow
\mathcal{O}_{t}
\rightarrow
\mathcal{P}_{t}
\rightarrow
\mathcal{C}_{t}
\rightarrow
\mathcal{E}_{t+1},
\label{eq:closed_loop}
\end{equation}

where $\mathcal{E}_{t}$ denotes the environment and robot state at time $t$, $\mathcal{O}_{t}$ denotes the observations transmitted to ROS2, $\mathcal{P}_{t}$ represents perception and planning, and $\mathcal{C}_{t}$ denotes the commands executed in Unity. The resulting state $\mathcal{E}_{t+1}$ provides the observations for the next simulation cycle.

This layered design separates the virtual environment from task-specific algorithms. Navigation, perception, or manipulation modules can therefore be replaced on the ROS2 side without reconstructing the Unity scene. Similarly, greenhouse layouts and sensor configurations can be modified without changing the overall planning and control interfaces.

\subsection{Virtual Greenhouse and Environmental Control}
\label{subsec:greenhouse_environment}

\begin{figure}[t]
    \centering
    \includegraphics[width=\columnwidth]
    {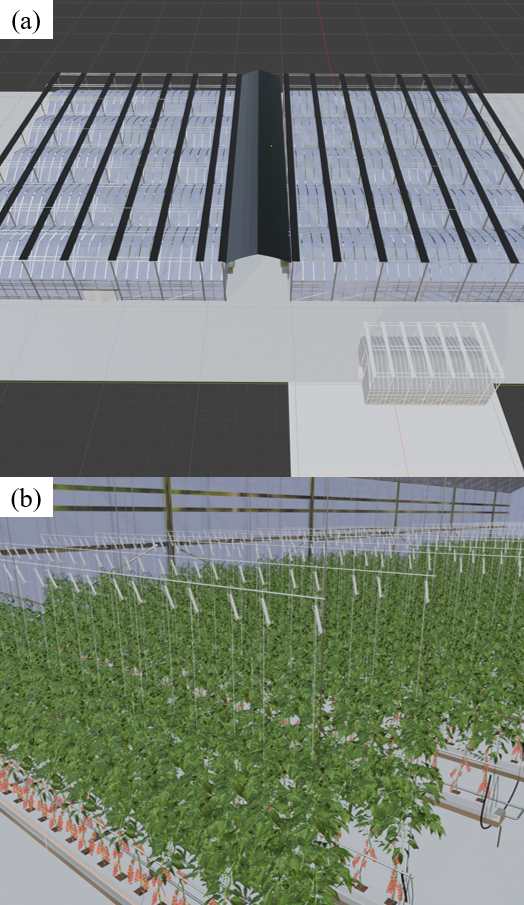}
    \caption{Construction of the virtual greenhouse:
    (a) greenhouse structural model and
    (b) crop-populated tomato-growing scene.}
    \label{fig:greenhouse_world}
\end{figure}

The greenhouse construction is illustrated in Fig.~\ref{fig:greenhouse_world}. The scene contains the greenhouse structure, densely arranged tomato rows, fruits, crop supports, cultivation facilities, ground surfaces, and navigable passages. Modular assets allow the greenhouse structure, plants, fruits, and obstacles to be rearranged without rebuilding the complete scene.

The crop rows and cultivation facilities form confined passages for mobile-robot navigation. Tomato fruits are distributed at different positions and orientations around the plants to provide manipulation targets. Their arrangement can be changed to create different levels of occlusion, target density, and manipulator accessibility. Obstacles can also be introduced into the passage or manipulator workspace to evaluate navigation safety and collision-aware motion planning.

\begin{figure}[t]
    \centering
    \includegraphics[width=0.76\columnwidth]
    {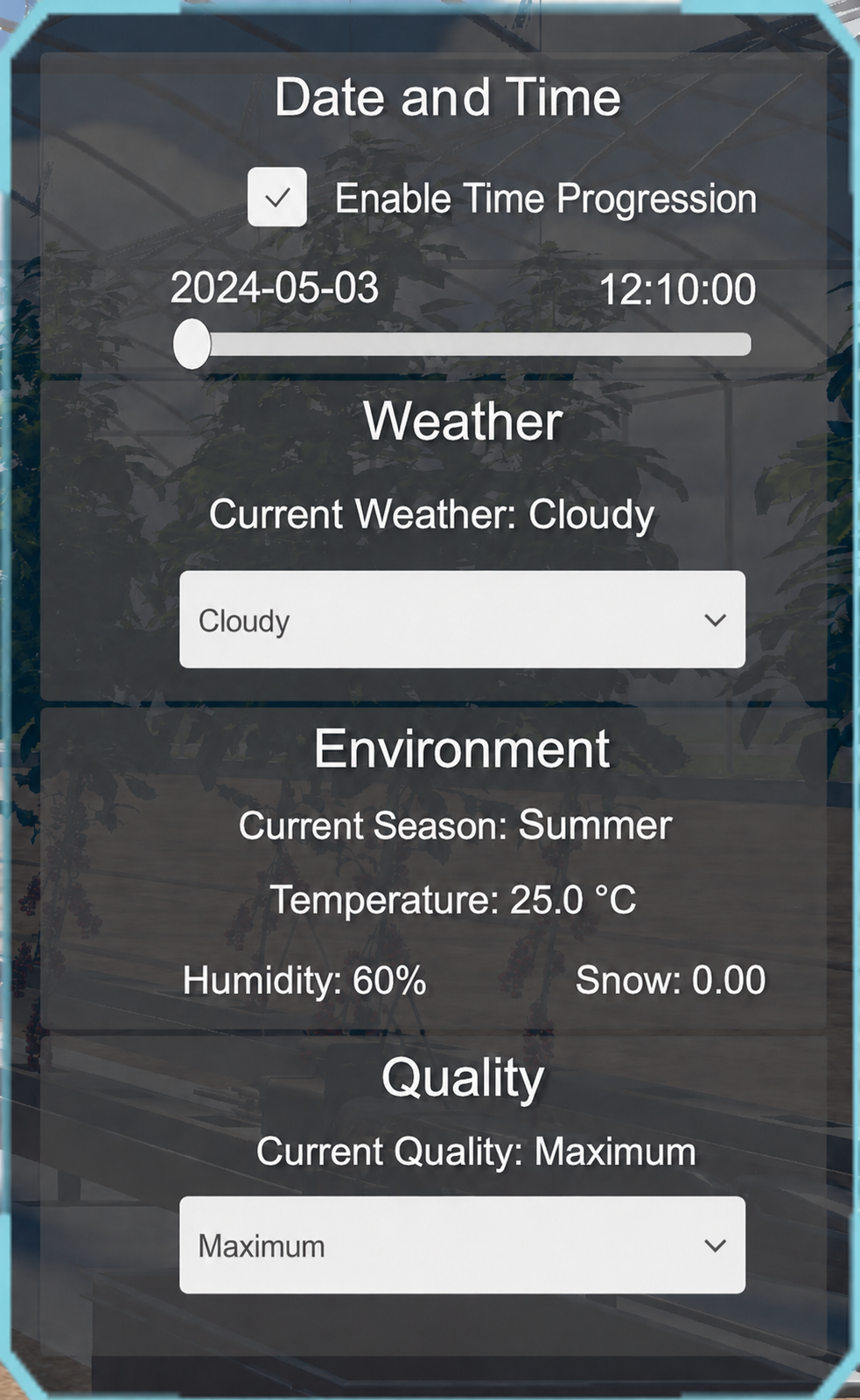}
    \caption{Environmental-control panel for configuring simulation date,
    time, weather, seasonal conditions, temperature, humidity, and
    rendering quality.}
    \label{fig:weather_panel}
\end{figure}

Agri-Sim provides an environmental-control panel for configuring date, time, weather, season, temperature, humidity, and rendering quality, as shown in Fig.~\ref{fig:weather_panel}. The date and time settings modify the solar direction and illumination condition of the greenhouse. Weather presets change scene appearance and atmospheric effects, while the rendering-quality setting provides a trade-off between visual fidelity and computational performance. These controls allow the same robot task to be evaluated under multiple environmental appearances without changing
the greenhouse geometry.

The scene configuration can be fixed before each experimental group to support repeatable evaluation. Unless an environmental factor is the independent variable, repeated trials use the same greenhouse layout, robot initial pose, target distribution, obstacle placement, sensor configuration, physics settings, and simulation-time parameters.

The current platform focuses on controllable geometric and visual conditions for navigation and motion-planning experiments. The plants and fruits use collision geometries for contact detection, but high-fidelity crop growth, flexible branch and leaf dynamics, and fruit-detachment mechanics are outside the present implementation.

\subsection{Robot Model and Virtual Sensors}
\label{subsec:robot_and_sensors}

\begin{figure}[t]
    \centering
    \includegraphics[width=0.96\columnwidth]
    {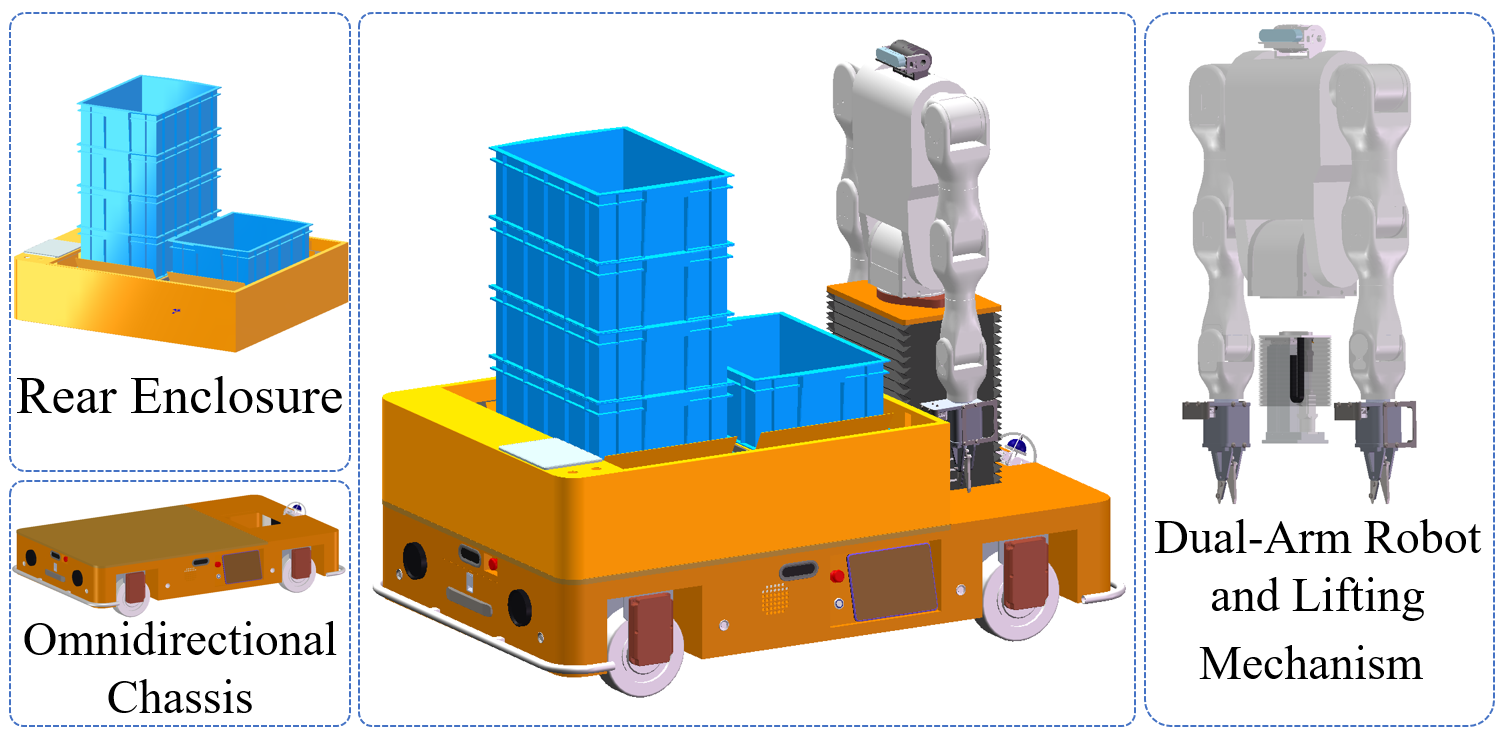}
    \caption{Mobile dual-arm harvesting robot and its principal modules:
    rear enclosure, omnidirectional chassis, dual-arm assembly, and
    lifting mechanism.}
    \label{fig:robot_model}
\end{figure}

The simulated harvesting robot consists of an omnidirectional mobile chassis, a lifting mechanism, two articulated manipulators, end effectors, and a rear equipment enclosure, as shown in
Fig.~\ref{fig:robot_model}. The mobile chassis provides planar motion through greenhouse passages, while the lifting mechanism extends the vertical workspace of the manipulators.

The Unity articulation hierarchy follows the joint names, axes, limits, and parent--child relationships defined in the robot description used by ROS2 and MoveIt~2. Maintaining the same kinematic structure in the two systems enables consistent joint-state feedback, inverse-kinematics
solutions, and trajectory execution.

The mobile base receives planar velocity commands

\begin{equation}
\mathbf{u}_{b}
=
\left[
v_{x},\,v_{y},\,\omega_{z}
\right]^{\mathrm{T}},
\label{eq:base_command}
\end{equation}

where $v_x$ and $v_y$ are the longitudinal and lateral velocities and $\omega_z$ is the yaw rate. The manipulators receive time-parameterized joint trajectories generated by MoveIt~2. Unity validates the joint names, trajectory dimensions, and temporal order before interpolating and executing the commanded joint positions.

\begin{figure}[t]
    \centering
    \includegraphics[width=\columnwidth]
    {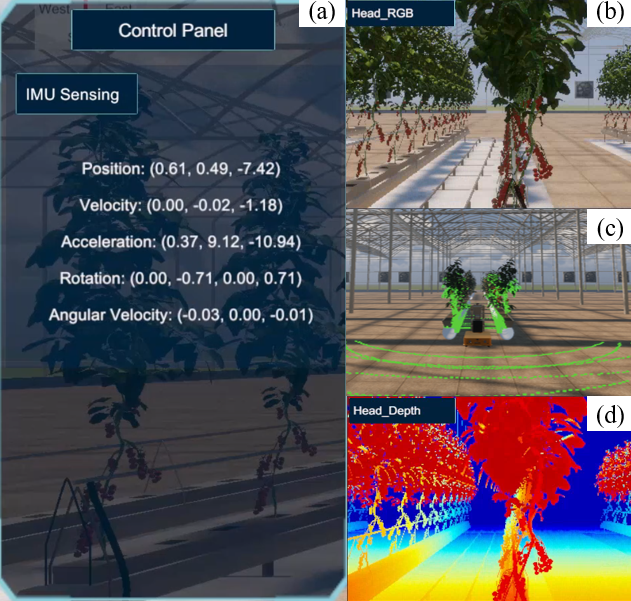}
    \caption{Representative virtual sensor outputs:
    (a) IMU measurements, (b) RGB observation,
    (c) LiDAR visualization, and (d) depth observation.}
    \label{fig:virtual_sensors}
\end{figure}

Agri-Sim provides RGB-D cameras, LiDAR, IMU, joint-state feedback, and collision-state monitoring. Representative outputs are shown in Fig.~\ref{fig:virtual_sensors}. Each virtual sensor has a configurable mounting pose, coordinate frame, acquisition rate, and ROS2 topic.

The RGB-D camera generates synchronized color and depth observations. Its configurable parameters include image resolution, field of view, clipping range, acquisition frequency, and mounting transformation. For a depth pixel $(u,v)$ with depth $d(u,v)$, the corresponding point in the camera optical frame is calculated as

\begin{equation}
\begin{split}
X &= \frac{(u-c_x)d(u,v)}{f_x},\\
Y &= \frac{(v-c_y)d(u,v)}{f_y},\\
Z &= d(u,v),
\end{split}
\label{eq:depth_projection}
\end{equation}

where $(f_x,f_y)$ are the focal lengths and $(c_x,c_y)$ is the principal point. Camera calibration information is transmitted with the images for standard ROS2 processing.

The virtual LiDAR uses ray-based surface intersection to generate three-dimensional points. Its horizontal and vertical fields of view, angular resolution, maximum range, scanning rate, and mounting pose can be configured. The generated point cloud can be used for obstacle perception, mapping, localization, and navigation.

The IMU publishes orientation, angular velocity, and linear acceleration in its local sensor frame. Joint-state messages provide joint names, positions, velocities, and available effort values so that MoveIt~2 can maintain a robot state consistent with Unity. Collision events contain the involved object identities and contact locations and are used to assess navigation safety and manipulation-path feasibility.

The current IMU implementation provides ideal or configurable virtual measurements. Detailed bias drift, scale-factor error, temperature dependence, and plant-contact deformation models may be introduced in future work when higher sensor or interaction fidelity is required.

\subsection{ROS2--Unity Communication}
\label{subsec:communication}

\begin{figure}[t]
    \centering
    \includegraphics[width=0.92\columnwidth]
    {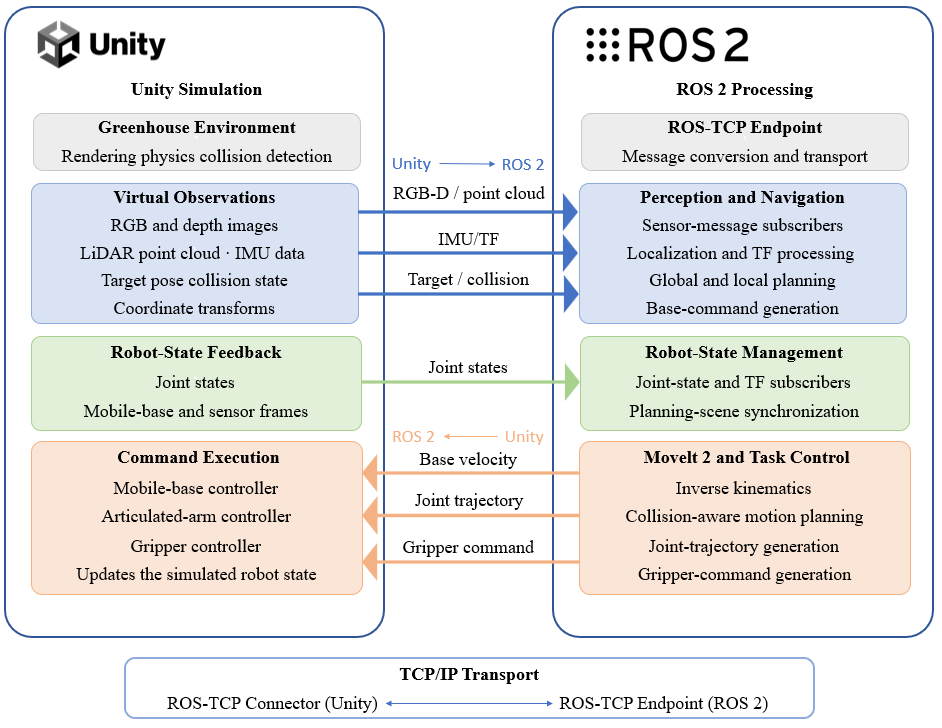}
    \caption{Bidirectional ROS2--Unity communication framework for virtual observations, target poses, robot-state feedback, and robot-motion commands.}
    \label{fig:ros_unity_communication}
\end{figure}

The bidirectional communication framework is shown in Fig.~\ref{fig:ros_unity_communication}. Unity publishes sensor observations, target poses, joint states, and collision information.
ROS2 processes these messages and returns mobile-base commands, manipulator trajectories, and gripper commands.

The principal communication interfaces are summarized in Table~\ref{tab:communication_interfaces}. Topic names, namespaces, message rates, and coordinate frames can be modified for different experiments.

\begin{table}[t]
\caption{Principal ROS2--Unity Communication Interfaces}
\label{tab:communication_interfaces}
\centering
\scriptsize
\setlength{\tabcolsep}{2.0pt}
\renewcommand{\arraystretch}{1.08}

\begin{tabularx}{\columnwidth}{
    >{\raggedright\arraybackslash}p{0.27\columnwidth}
    >{\centering\arraybackslash}p{0.24\columnwidth}
    >{\raggedright\arraybackslash}X}
\hline
\textbf{Data} &
\textbf{Direction} &
\textbf{ROS2 message type} \\
\hline

RGB/depth images &
Unity $\rightarrow$ ROS2 &
\makecell[l]{\texttt{sensor\_msgs/msg/}\\\texttt{Image}} \\

Camera parameters &
Unity $\rightarrow$ ROS2 &
\makecell[l]{\texttt{sensor\_msgs/msg/}\\\texttt{CameraInfo}} \\

Point cloud &
Unity $\rightarrow$ ROS2 &
\makecell[l]{\texttt{sensor\_msgs/msg/}\\\texttt{PointCloud2}} \\

IMU data &
Unity $\rightarrow$ ROS2 &
\makecell[l]{\texttt{sensor\_msgs/msg/}\\\texttt{Imu}} \\

Joint states &
Unity $\rightarrow$ ROS2 &
\makecell[l]{\texttt{sensor\_msgs/msg/}\\\texttt{JointState}} \\

Coordinate transforms &
Unity $\rightarrow$ ROS2 &
\makecell[l]{\texttt{tf2\_msgs/msg/}\\\texttt{TFMessage}} \\

Target pose &
Unity $\rightarrow$ ROS2 &
\makecell[l]{\texttt{geometry\_msgs/msg/}\\\texttt{PoseStamped}} \\

Base command &
ROS2 $\rightarrow$ Unity &
\makecell[l]{\texttt{geometry\_msgs/msg/}\\\texttt{Twist}} \\

Joint trajectory &
ROS2 $\rightarrow$ Unity &
\makecell[l]{\texttt{trajectory\_msgs/msg/}\\\texttt{JointTrajectory}} \\

Gripper command &
ROS2 $\rightarrow$ Unity &
Custom/configurable message \\

Collision state &
Unity $\rightarrow$ ROS2 &
Custom/configurable message \\
\hline
\end{tabularx}
\end{table}

The communication implementation retains the most recent valid base command instead of accumulating outdated commands. Manipulator trajectories are checked for matching joint names, valid point dimensions, and monotonically increasing timestamps. Invalid commands are rejected and reported before being applied to the simulated robot.

\subsection{Coordinate and Time Consistency}
\label{subsec:coordinate_time}

Unity and ROS2 use different default coordinate conventions. Unity commonly uses a left-handed coordinate system, whereas ROS2 follows right-handed coordinate conventions. Agri-Sim therefore applies explicit conversion to positions, orientations, linear and angular velocities, target poses, and collision-contact coordinates.

Let $\mathbf{p}_{U}$ be a position in the Unity frame and
$\mathbf{p}_{R}$ the corresponding position in the ROS2 frame. Their
relationship is

\begin{equation}
\mathbf{p}_{R}
=
\mathbf{C}_{UR}\mathbf{p}_{U},
\label{eq:coordinate_conversion}
\end{equation}

where $\mathbf{C}_{UR}$ is the Unity-to-ROS2 axis-conversion matrix.
In the implemented convention, the Unity axes (right, up, and forward)
are mapped to the ROS2 axes (forward, left, and up), giving
$x_R=z_U$, $y_R=-x_U$, and $z_R=y_U$. The coordinate hierarchy contains
the world or map frame, mobile-base frame, manipulator-link frames,
end-effector frames, and sensor frames. Fixed sensor mountings are
represented by static transforms, while the mobile base and articulated
joints generate dynamic transforms.

Target poses are published with timestamps and frame identifiers and are transformed into the corresponding MoveIt~2 planning frame before motion planning. Frame-visualization and pose-debugging components in Unity are used together with ROS2 TF tools to verify target, collision, link, and sensor-frame consistency.

All virtual sensors and robot-state interfaces use a common simulation clock. Sensor acquisition is scheduled according to simulation time instead of being determined directly by the rendering frame rate. A sensor with acquisition frequency $f_s$ is triggered at the simulated-time interval

\begin{equation}
T_s = \frac{1}{f_s}.
\label{eq:sensor_period}
\end{equation}

This scheduling strategy preserves the temporal relationship among RGB, depth, LiDAR, IMU, joint-state, and coordinate-transform messages even when the rendering rate changes. When ROS2 nodes use simulated time, the Unity clock is also published to the ROS2 system.

\subsection{Integrated Task Workflows}
\label{subsec:workflows}

Agri-Sim supports navigation and manipulation within the same greenhouse, robot model, communication framework, and simulation clock.

During navigation, Unity publishes LiDAR or RGB-D observations, IMU measurements, robot states, and coordinate transforms. ROS2 performs localization, perception, and path planning and returns a velocity command to the mobile base. Unity executes the command, updates the robot pose, checks for collisions, and publishes the next observations. The resulting loop is

\begin{equation}
\begin{aligned}
\text{sensing}
&\rightarrow
\text{localization and planning} \\
&\rightarrow
\text{base control}
\rightarrow
\text{Unity execution}.
\end{aligned}
\label{eq:navigation_loop}
\end{equation}

During manipulation, Unity publishes a tomato target pose in a specified coordinate frame. ROS2 transforms the target into the MoveIt~2 planning frame, after which MoveIt~2 performs inverse kinematics, collision checking, and sampling-based motion planning through the OMPL planning framework \cite{sucan2012ompl}. A valid time-parameterized joint trajectory is returned to Unity and executed by the articulated robot. Updated joint states and collision events are continuously published during execution. The manipulation loop is summarized as

\begin{equation}
\begin{aligned}
\text{target pose}
&\rightarrow
\text{frame transformation and planning} \\
&\rightarrow
\text{joint trajectory}
\rightarrow
\text{Unity execution}.
\end{aligned}
\label{eq:manipulation_loop}
\end{equation}

The current benchmark evaluates target-pose transmission, coordinate transformation, collision-aware planning, trajectory communication, single-arm tomato acquisition, inter-arm handover, box placement, and closed-loop mobile navigation. Autonomous fruit detection, physically based fruit detachment, high-fidelity plant deformation, force-controlled manipulation, and Sim-to-Real transfer remain outside the scope of the current experimental validation.

\section{Experimental Evaluation}
\label{sec:experimental_evaluation}

This section evaluates the functional effectiveness of Agri-Sim through ROS2--Unity communication tests, autonomous greenhouse navigation, and dual-arm tomato-harvesting experiments. The experiments were designed to verify whether the proposed platform could support closed-loop interaction between simulated sensors and external ROS2 algorithms, rather than to benchmark a particular navigation or manipulation algorithm.

\subsection{Experimental Setup}
\label{subsec:experimental_setup}

The Unity simulation was executed on a workstation equipped with an Intel Core i7-14700KF processor, NVIDIA GeForce RTX~5090 graphics card, and Windows~11 Pro. The ROS2 system was deployed on a separate Ubuntu~22.04 workstation equipped with an NVIDIA GeForce RTX~3080~Ti GPU and ROS2 Humble. The two computers were connected through the same local-area network.

The complete virtual greenhouse measured approximately
$70 \times 55 \times 10~\mathrm{m}$ and contained 160 configurable crop rows. To maintain a stable computational load during the reported experiments, six crop rows were enabled. The width of each greenhouse aisle was approximately 2~m, providing sufficient space for navigation and manipulation by the mobile dual-arm harvesting robot.

The simulated robot consisted of a four-wheel-steering omnidirectional mobile base, a vertical lifting mechanism, and two 7-DoF manipulators. The two manipulators therefore provided 14 primary arm degrees of freedom, excluding the grippers, lifting mechanism, waist, and mobile-base joints. The robot model was imported from its URDF description so that the link hierarchy, joint limits, collision geometry, and inertial properties could be shared with the ROS2 planning environment.

Unity used a fixed physics timestep of

\begin{equation}
\Delta t_{\mathrm{physics}} = 0.02~\mathrm{s},
\end{equation}

corresponding to a nominal physics-update frequency of

\begin{equation}
f_{\mathrm{physics}}
=
\frac{1}{\Delta t_{\mathrm{physics}}}
=
50~\mathrm{Hz}.
\end{equation}

The principal experimental configuration is summarized in
Table~\ref{tab:experimental_configuration}.

\begin{table}[t]
\centering
\caption{Principal experimental configuration.}
\label{tab:experimental_configuration}
\begin{tabular}{ll}
\hline
\textbf{Item} & \textbf{Configuration} \\
\hline
Unity operating system & Windows 11 Pro \\
Unity CPU & Intel Core i7-14700KF \\
Unity GPU & NVIDIA RTX 5090 GPU \\
ROS operating system & Ubuntu 22.04 \\
ROS distribution & ROS2 Humble \\
ROS GPU & NVIDIA RTX 3080 Ti \\
Network connection & Same local-area network \\
Greenhouse size & $70 \times 55 \times 10~\mathrm{m}$ \\
Complete crop layout & 160 rows \\
Rows enabled in experiments & 6 rows \\
Greenhouse aisle width & 2 m \\
Manipulator configuration & Dual 7-DoF arms \\
Unity fixed timestep & 0.02 s \\
Nominal physics frequency & 50 Hz \\
Unity--ROS publication rate & Configured at 10 Hz \\
ROS--Unity publication rate & Configured at 10 Hz \\
\hline
\end{tabular}
\end{table}

\subsection{Virtual Sensor Configuration}
\label{subsec:sensor_configuration}

The simulated robot was equipped with an RGB-D camera, a virtual LiDAR, an IMU, and joint-state sensors. These sensors generated the perception and robot state information required by the ROS2 navigation and manipulation modules.

The virtual RGB-D camera was configured using selected parameters
derived from an Intel RealSense D435i profile. The depth image resolution was $1280 \times 720$ pixels, and the sensor output was configured for publication at 10~Hz. The vertical field of view was set to $58^{\circ}$, while the corresponding horizontal field of view was approximately $87^{\circ}$. The simulated depth range was 0.105--10~m. Zero-mean Gaussian noise with a standard-deviation parameter of $0.015~\mathrm{m}$ was applied to the generated depth data to avoid completely ideal depth measurements.

The virtual LiDAR was implemented using a general-purpose ray-casting sensor configured according to the principal range, field-of-view, and scanning parameters of the Livox MID-360 used on the physical robot. Therefore, the sensor provided a MID-360-compatible parameter configuration, but it did not fully reproduce the proprietary non-repetitive scanning pattern of the physical device.

The generated RGB-D images, point clouds, IMU measurements, robot transforms, and joint states were transmitted from Unity to ROS2. Navigation commands and manipulator joint-trajectory commands were transmitted in the reverse direction. The communication frequencies could be adjusted according to the requirements of different topics. For the experiments reported in this work, both information publication and reception were configured at 10~Hz.

\subsection{Evaluation Criteria}
\label{subsec:evaluation_criteria}

Three aspects of the platform were evaluated:

\begin{enumerate}
    \item whether Unity and ROS2 could continuously exchange sensor,
    state, and command messages;
    \item whether the mobile robot could reach a specified greenhouse
    waypoint without collision; and
    \item whether the dual-arm robot could complete the entire tomato-
    harvesting sequence.
\end{enumerate}

No end-to-end, one-way, or round-trip communication latency was measured in this study. Consequently, the evaluation was limited to functional message
exchange and continuity of closed-loop task execution, rather than
quantitative communication performance.

For the navigation experiment, the goal region was defined as a circle with a fixed radius of $250~\mathrm{mm}$ centered at the target waypoint. This radius was specified independently of the robot length. A trial was considered successful only when the geometric center of the mobile base entered this region without collision within $120~\mathrm{s}$. A trial was unsuccessful if the robot collided with an obstacle or did not enter the goal region before the time limit. Navigation was evaluated only as a binary success or failure; positioning-error magnitudes were not used as an evaluation metric.

For the harvesting experiment, a trial was considered successful only when the robot completed all of the following stages:

\begin{enumerate}
    \item generation of executable manipulator trajectories;
    \item approach and grasping of the target tomato by the first arm;
    \item attachment of the tomato to the gripper in Unity;
    \item transfer of the tomato between the two manipulators; and
    \item placement of the tomato into the harvesting box.
\end{enumerate}

The maximum planning time for each MoveIt~2 motion-planning request was set to 15~s. A target attachment event was used to determine whether the initial tomato grasp had been established successfully.

\subsection{ROS2--Unity Closed-Loop Functional Verification}
\label{subsec:communication_test}

The communication test verified the bidirectional transmission of virtual sensor data, robot states, and control commands. Unity continuously published RGB-D data, virtual LiDAR point clouds, IMU data, TF transformations, and joint states to ROS2. ROS2 received these messages and used them as inputs to the navigation and MoveIt~2 planning modules.

In the reverse direction, ROS2 published mobile-base control commands and manipulator trajectory commands to Unity. Unity received these commands and applied them to the corresponding Articulation Body components of the robot model.

Both message publication and reception were configured at 10~Hz during the reported experiments. The communication frequencies refer to the transfer of new messages, rather than the internal execution-loop frequency of the corresponding Unity or ROS2 nodes. The bidirectional
connection remained operational during the navigation and tomato-harvesting experiments, allowing the complete perception--planning--control loop to be executed without manual transfer of intermediate data.

Because no dedicated timestamp-based latency experiment was performed, communication delay and round-trip latency are not reported. The results in this subsection demonstrate functional closed-loop connectivity but should not be interpreted as a quantitative network-latency benchmark.

\subsection{Autonomous Navigation Evaluation}
\label{subsec:navigation_evaluation}

The navigation experiment evaluated whether the mobile robot could safely reach a specified target waypoint in the greenhouse aisle. A classical ROS2 navigation architecture was adopted, following the modular organization of contemporary Navigation2 systems \cite{macenski2020marathon2}. The A* algorithm \cite{hart1968astar} was used for global path planning, while the Dynamic Window Approach (DWA) \cite{fox1997dwa} was used for local trajectory generation, obstacle avoidance, and velocity control. Global and local cost maps were used to represent the greenhouse boundaries, crop rows,
and nearby obstacles.

For each trial, the target waypoint was provided to the ROS2 navigation module. The planned velocity commands were transmitted to Unity at the configured rate of 10~Hz and applied to the simulated mobile base. The goal region was a circle with a fixed radius of $250~\mathrm{mm}$ centered at the target waypoint. A trial was classified as successful if the geometric center of the mobile base entered this region without collision within $120~\mathrm{s}$. A collision or failure to enter the region within $120~\mathrm{s}$ was recorded as an unsuccessful trial. No distance-error bins or other positioning-error statistics were used in the navigation evaluation.

\begin{figure*}[t]
    \centering
    \includegraphics[width=0.98\textwidth]
    {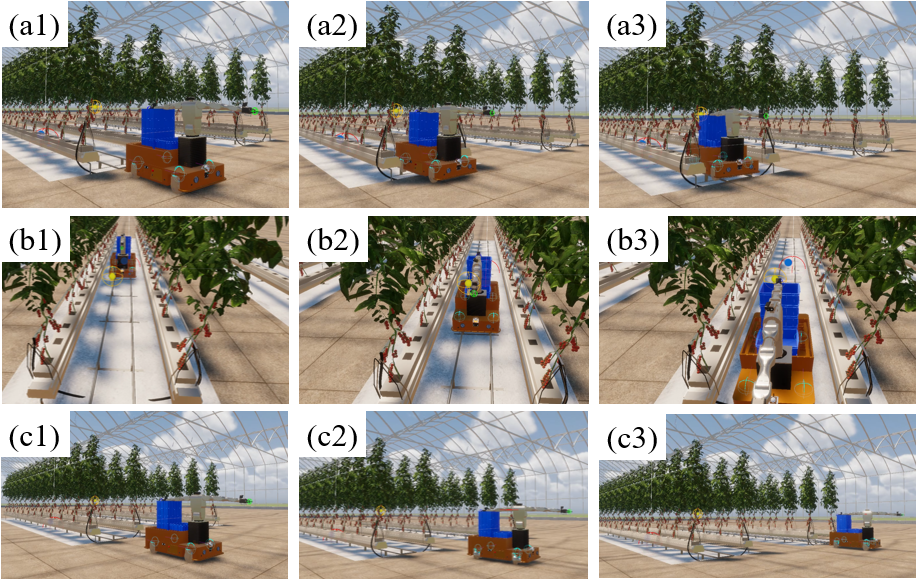}
    \caption{Representative frames of the autonomous greenhouse-navigation sequence: (a1)--(a3) the mobile robot enters the crop-row track; (b1)--(b3) the robot traverses along the track between two crop rows; and (c1)--(c3) the robot performs a row-switching maneuver to enter the adjacent track.}
    \label{fig:navigation_sequence}
\end{figure*}

Representative frames from a successful navigation trial are shown in Fig.~\ref{fig:navigation_sequence}. The sequence illustrates the mobile robot moving through the greenhouse aisle while maintaining clearance from the crop rows and cultivation facilities.

A total of 50 navigation trials were performed using different target
waypoints under the same greenhouse layout, obstacle configuration,
sensor configuration, and predefined initial robot state. Before each trial, the robot was reset to the predefined initial state. Among the 50 trials, the robot satisfied the predefined goal-region criterion without collision in 45 trials. The remaining five trials were classified as unsuccessful because they did not satisfy the 250-mm-radius goal-region criterion. Detailed trial-level records were not retained, and the five failures were therefore not further categorized. The navigation success rate was calculated as

\begin{equation}
R_{\mathrm{nav}}
=
\frac{N_{\mathrm{reached}}}{N_{\mathrm{total}}}
\times 100\%
=
\frac{45}{50}\times100\%
=
90\%.
\end{equation}

Using the Wilson score interval for a binomial proportion, the corresponding
95\% confidence interval (CI) was 78.6\%--95.7\%. The navigation results are
summarized in Table~\ref{tab:navigation_results}.

\begin{table}[t]
\centering
\caption{Autonomous navigation results over 50 repeated trials.}
\label{tab:navigation_results}
\begin{tabular}{lccc}
\hline
\textbf{Navigation outcome} & \textbf{Trials} & \textbf{Rate} & \textbf{95\% CI} \\
\hline
Success & 45 & 90\% & 78.6--95.7\% \\
Failure & 5 & 10\% & --- \\
\hline
Total & 50 & 100\% & --- \\
\hline
\end{tabular}
\end{table}

These results indicate that the ROS2 navigation stack could control the virtual robot through the Unity--ROS2 interface and complete the greenhouse navigation task in most repeated trials. Historical trial-by-trial records were not retained. Therefore, only the aggregate binary outcome is reported; navigation time, path length, and detailed failure trajectories cannot be reconstructed retrospectively.

\subsection{Tomato-Harvesting Evaluation}
\label{subsec:harvesting_evaluation}

The manipulation experiment evaluated the complete harvesting workflow rather than only the initial grasp. The target tomato pose was obtained directly from the Unity scene and published to ROS2 as a ground-truth target pose. Therefore, the experiment evaluated coordinate transformation, motion planning, and task execution rather than autonomous fruit detection or pose estimation. MoveIt~2 generated collision-aware trajectories for the manipulators, and the resulting joint commands were sent to Unity at the configured rate of 10~Hz.

The first manipulator approached and grasped the selected tomato. Once the target was attached to the gripper, the two manipulators performed an inter-arm handover. The second manipulator subsequently moved the tomato to the harvesting box and released it. The complete sequence was considered successful only if the tomato was grasped, transferred, and placed into the box.

\begin{figure*}[t]
    \centering
    \includegraphics[width=0.98\textwidth]
    {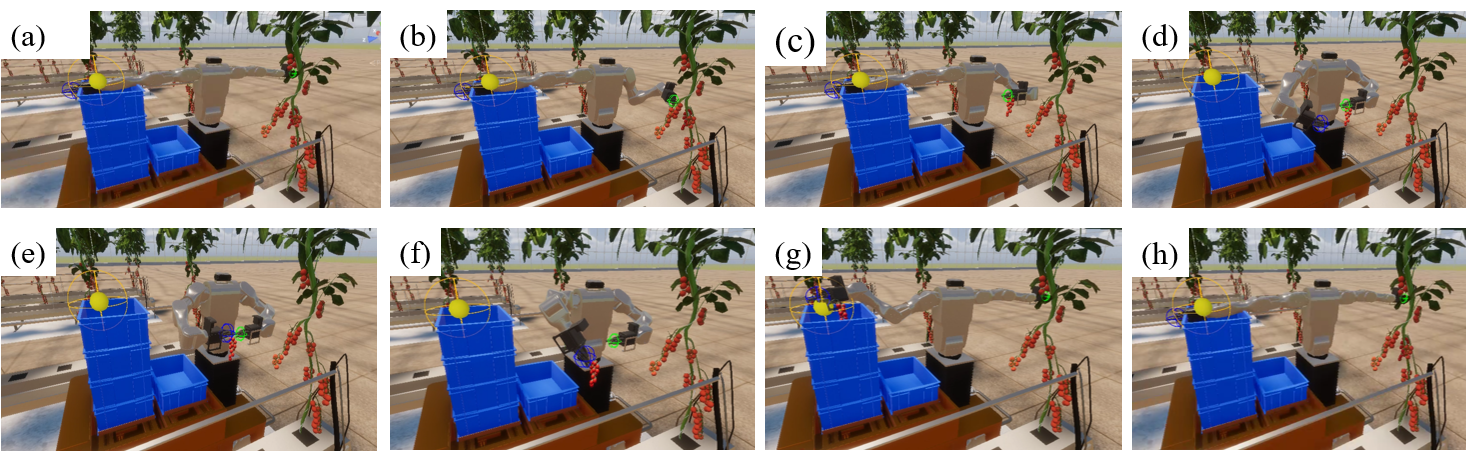}
    \caption{Representative frames of the dual-arm tomato-harvesting workflow: (a) target selection; (b) pre-grasp positioning; (c) tomato grasping; (d) preparation for inter-arm handover; (e) inter-arm handover; (f) transfer toward the harvesting box; (g) placement into the box; and (h) task completion.}
    \label{fig:manipulation_sequence}
\end{figure*}

Representative frames of the complete manipulation workflow are presented in Fig.~\ref{fig:manipulation_sequence}. The sequence includes target approach, tomato acquisition by the first manipulator, inter-arm handover, and final placement into the harvesting box.

A total of 50 repeated tomato-harvesting trials were conducted. A trial was classified as planning-successful only when executable trajectories were generated for all motion-planning requests required by the acquisition, handover, and placement stages. Executable motion trajectories were successfully generated in 48 trials, corresponding to a motion-planning success rate of

\begin{equation}
R_{\mathrm{planning}}
=
\frac{48}{50}\times100\%
=
96\%.
\end{equation}

Among the 48 trials with successful motion planning, 43 completed the entire harvesting sequence. The overall harvesting success rate was therefore

\begin{equation}
R_{\mathrm{harvesting}}
=
\frac{43}{50}\times100\%
=
86\%.
\end{equation}

The conditional harvesting success rate after successful motion planning was

\begin{equation}
R_{\mathrm{harvesting|planning}}
=
\frac{43}{48}\times100\%
\approx
89.6\%.
\end{equation}

The Wilson 95\% CIs were 86.5\%--98.9\% for motion-planning success and
73.8\%--93.1\% for complete-task success. The detailed outcomes are shown in
Table~\ref{tab:harvesting_results}.

\begin{table}[t]
\centering
\caption{Tomato-harvesting results over 50 repeated trials.}
\label{tab:harvesting_results}
\begin{tabular}{lcc}
\hline
\textbf{Outcome} & \textbf{Trials} & \textbf{Rate} \\
\hline
Complete harvesting sequence successful & 43 & 86\% \\
Initial grasp failure & 2 & 4\% \\
Inter-arm handover failure & 2 & 4\% \\
Box-placement failure & 1 & 2\% \\
Motion-planning failure & 2 & 4\% \\
\hline
Total & 50 & 100\% \\
\hline
\end{tabular}
\end{table}

Of the five trials in which motion planning succeeded but the complete harvesting task failed, two failures occurred because the first manipulator did not successfully grasp and attach the tomato. Two failures occurred during the inter-arm handover, and one failure occurred while placing the tomato into the harvesting box.

The remaining two trials failed during the motion-planning stage. In these cases, the planner either exceeded the 15~s planning-time limit or failed to generate a valid trajectory. Since detailed planner logs were not retained, the two planning failures were not further divided into timeout and trajectory-generation categories.

These results show that the platform supported not only collision-aware single-arm motion planning but also coordinated dual-arm task execution, including target attachment, handover, and final placement.

\subsection{Simulation-Time and Runtime Performance}
\label{subsec:runtime_performance}

The Unity physics engine operated with a fixed timestep of 0.02~s, giving a nominal physics-update frequency of 50~Hz. This update frequency was higher than the configured 10~Hz communication frequency used for bidirectional Unity--ROS2 message transfer. Consequently, several physics updates could occur between two consecutive control or sensor messages.

The relationship between the physics and communication updates was

\begin{equation}
N_{\mathrm{physics/message}}
=
\frac{f_{\mathrm{physics}}}{f_{\mathrm{communication}}}
=
\frac{50}{10}
=
5,
\end{equation}

meaning that approximately five Unity physics steps were executed during each communication interval.

The navigation and harvesting experiments were completed with the RGB-D camera, LiDAR, IMU, joint-state feedback, and TF publication enabled. No dedicated frame-rate logging experiment was performed, and therefore an exact mean FPS, minimum FPS, or frame-time distribution is not reported. The reported experiments only confirm that the six-row greenhouse configuration could maintain functional task execution and continuous ROS2--Unity message exchange on the specified hardware.

The RTX~5090 GPU describes the hardware available on the Unity workstation; no dedicated GPU-utilization or rendering-performance measurements were conducted in this study.

\subsection{Discussion of Experimental Results}
\label{subsec:experimental_discussion}

The experiments verify the intended closed-loop workflow rather than the superiority of a navigation or manipulation algorithm. The difference between the 96\% planning success rate and the 86\% complete-task success rate shows why trajectory generation and full dual-arm execution must be reported separately: attachment, handover, release, and box placement can fail even
after all required trajectories have been generated.

\section{Discussion}
\label{sec:discussion}

Agri-Sim provides modular closed-loop integration between a Unity greenhouse and external ROS2 navigation and manipulation modules. The experiments confirm the consistency of its communication interfaces, coordinate frames, planning scene, and command execution for greenhouse mobile manipulation.

The fixed $250~\mathrm{mm}$ navigation goal radius and $120~\mathrm{s}$ time limit provide an explicit binary task definition. The resulting 45/50 outcome supports functional integration but should not be interpreted as a precise localization benchmark, because positioning-error magnitudes were intentionally excluded and historical per-trial trajectories were not retained. Likewise, the harvesting results assess the logical and kinematic execution of the full dual-arm workflow; tomato attachment in Unity does not reproduce gripping force, fruit deformation, stem detachment, or damage.

Several limitations remain. The crop models primarily reproduce visual appearance and collision geometry; deformable behavior, fruit damage, force-dependent detachment, and detailed grasp-contact mechanics are not yet represented.

The virtual sensors approximate their physical counterparts: the ray-casting LiDAR does not reproduce the MID-360's complete non-repetitive scan pattern, and the D435i profile omits effects such as motion blur, reflective-surface artifacts, and missing depth measurements.

Communication was functionally verified at the configured 10~Hz against a 50~Hz physics update, but latency, jitter, bandwidth, and packet loss were not measured; it is therefore not a network-performance benchmark.

Finally, experiments used six active crop rows; the 160-row scene was not evaluated for frame rate, real-time factor, memory, or communication load. Only aggregated outcomes were retained, and physical-robot experiments were outside the scope; thus, detailed trial-level statistics and Sim-to-Real performance remain unestablished.

\section{Conclusion}
\label{sec:conclusion}

This paper presented Agri-Sim, a Unity--ROS2 platform integrating a configurable greenhouse, a mobile dual-arm harvesting robot, multimodal virtual sensors, and bidirectional execution of ROS2 navigation commands and MoveIt~2-generated trajectories.

In 50 trials per task, navigation succeeded in 45 cases (90\%; Wilson 95\% CI: 78.6\%--95.7\%), trajectory planning in 48 (96\%; 86.5\%--98.9\%), and the complete dual-arm harvesting workflow in 43 (86\%; 73.8\%--93.1\%). These results demonstrate closed-loop integration of virtual sensing, navigation, collision-aware planning, and coordinated dual-arm execution.

The evaluation verifies functionality rather than algorithmic superiority. Future work will add deformable crop and force-aware contact models, calibrate virtual sensors, retain trial-level data, and evaluate communication, runtime, and scalability. Domain randomization, parallel learning interfaces, and physical-robot experiments will support quantitative Sim-to-Real evaluation.

\section*{Data and Code Availability}

The source code, Unity scenes, ROS2 interface definitions, and detailed experimental files are not publicly available at present. Reasonable requests for research access may be directed to the corresponding authors.

\section*{Acknowledgements}

This work was supported by the Shanghai Municipal Commission of Agriculture and Rural Affairs under Grant No.~T2025301.

\section*{Declaration of Competing Interest}

The authors declare that they have no known competing financial interests or personal relationships that could have appeared to influence the work reported in this paper.

\section*{CRediT Authorship Contribution Statement}

Shuhan Shi: Methodology, Software, Investigation, Visualization, Writing -- original draft, Writing -- review \& editing.
Zhenfeng Xue: Conceptualization, Methodology, Supervision.
Minghao Mei: Investigation.
Chao Zheng: Visualization; Methodology.
Nan Li: Visualization; Methodology; Software.
Zhonghua Miao: Conceptualization, Project administration, Funding acquisition.

\bibliographystyle{elsarticle-num}
\bibliography{refs}

\end{document}